\documentclass{article}

\usepackage{arxiv}

\usepackage[utf8]{inputenc} 
\usepackage[T1]{fontenc}    
\usepackage{hyperref}       
\usepackage{url}            
\usepackage{booktabs}       
\usepackage{amsfonts}       
\usepackage{nicefrac}       
\usepackage{microtype}      
\usepackage{lipsum}		
\usepackage{graphicx}
\usepackage{natbib}
\usepackage{doi}

\usepackage[linesnumbered,ruled,vlined]{algorithm2e}
\usepackage{mathtools}

\title{FedBot: Enhancing Privacy in Chatbots with Federated Learning}

\author{Addi Ait-Mlouk \\
        School of Informatics, University of Sk\"ovde,\\
        Sk\"ovde, Sweden\\
        \texttt{addi.ait-mlouk@his.se \thanks{Corresponding author.}}
         \And
        Sadi Alawadi\\
        School of Information Technology\\
        Halmstad University, Halmstad, Sweden\\
        \texttt{sadi.alawadi@hh.se}
        \And
        Salman Toor\\
        Department of Information Technology\\
        Division of Scientific Computing\\
        Uppsala University, Sweden\\
        Scaleout Systems, Sweden\\
        \texttt{salman.toor@it.uu.se}
         \And
         Andreas Hellander \\
        Department of Information Technology\\
        Division of Scientific Computing\\
        Uppsala University, Sweden\\
        Scaleout Systems, Sweden\\
        \texttt{andreas.hellander@it.uu.se}
}

\renewcommand{\shorttitle}{FedBot: Enhancing Privacy in Chatbots with Federated Learning}

\hypersetup{
pdftitle={A template for the arxiv style},
pdfsubject={q-bio.NC, q-bio.QM},
pdfauthor={David S.~Hippocampus, Elias D.~Striatum},
pdfkeywords={First keyword, Second keyword, More},
}

\begin{document}
\maketitle

\begin{abstract}
Chatbots are mainly data-driven and usually based on utterances that might be sensitive. However, training deep learning models on shared data can violate user privacy. Such issues have commonly existed in chatbots since their inception. In the literature, there have been many approaches to deal with privacy, such as differential privacy and secure multi-party computation, but most of them need to have access to users' data. In this context, Federated Learning (FL) aims to protect data privacy through distributed learning methods that keep the data in its location. This paper presents Fedbot, a proof-of-concept (POC) privacy-preserving chatbot that leverages large-scale customer support data. The POC combines Deep Bidirectional Transformer models and federated learning algorithms to protect customer data privacy during collaborative model training. The results of the proof-of-concept showcase the potential for privacy-preserving chatbots to transform the customer support industry by delivering personalized and efficient customer service that meets data privacy regulations and legal requirements. Furthermore, the system is specifically designed to improve its performance and accuracy over time by leveraging its ability to learn from previous interactions.

\end{abstract}


\maketitle
\section{Background}
A chatbot is a virtual agent able to assist users by providing useful information to the user's request. Chatbots have been very popular since their inception in 1960. After two decades, research and development have seen impressive progress from Eliza in 1960 to AI chatbots such as Siri in 2010, Cortona, and google assistant. Early chatbot systems, such as Eliza \cite{Weizenbaum:1966:ECP:365153.365168}, Parry \cite{Colby1999}, and Alice \cite{alice001}, was designed based on a text conversation. These chatbots have been used in Healthcare, e-commerce, and marketing. They are not just conversational systems but also can carry out other tasks such as ordering, booking, frequently asked questions, and symptom assessment. In the past several years, large companies have invested in AI and developed several assistants, among them Apple's Siri \footnote{https://www.apple.com/ios/siri/}, Microsoft Cortana \footnote{https://www.microsoft.com/en-us/cortana/}, Google Assistant \footnote{https://assistant.google.com/}, Facebook Messenger \footnote{https://developers. Facebook.com/blog/post/2016/04/12/}, Alexa \footnote{https://developer.amazon.com/alexa/} and ChatGPT \footnote{https://chat.openai.com/}. In terms of security, fundamental challenges  for chatbots and Question answering systems are trustworthiness and data privacy. There are concerns regarding the proper handling of data in terms of consent and regulations; these concerns revolve around 1) where and how data is shared with third parties, 2) how data is legally collected or stored, 3) how data is regulated (GDPR, HIPAA, GLBA, CCPA).\\

Chatbots are mainly data-driven; training a chatbot model requires massive training data. In most machine learning settings, users or companies must upload their data to a central server in order to train chatbot models. However, centralizing data in this way is not always possible. Reasons for this include the data being large, sensitive, and legally regulated. Federated learning offers several benefits for chatbot applications. First, it allows for the training of chatbots on a large dataset without the need to transfer all the data to a central server which is particularly useful for chatbot applications that handle sensitive or private information. Second, federated learning enables the training of chatbots on data from multiple sources, which can improve the accuracy and robustness of the model, this is especially important for chatbots that interact with diverse user populations. Third, federated learning can reduce the computational burden on the central server, since the training takes place on local devices to ensure that user privacy is preserved, as the data never leaves the device. In recent work, we developed a flexible and scalable framework for federated learning, FEDn, and demonstrated it for a  wide range of applications in a geographically distributed setting ~\cite{ekmefjord2021scalable}. To extend it to more sophisticated applications, firstly, we propose in this paper our POC that leverages large customer support datasets to train collaboratively Deep Bidirectional Transformer models for customer chatbots.
Secondly, we employed incremental online learning to construct more demanding models that enable accurate local training. This method enhances model performance during collaborative training without sacrificing data privacy. It also enables cross-silo and cross-device participation, allowing models to learn from distributed datasets while sharing knowledge in a trusted environment.

The remainder of this paper is organized as follows. Section \ref{sec:relatedwork} surveys related work. Section \ref{sec:approach} details the proposed approach and architecture of Fedbot, with an emphasis on its privacy and scalability properties. In Section \ref{sec:experiments}, we demonstrate the potential of the framework in an evaluation based on customer support datasets. Finally, Section 4  concludes the work and outlines future work.

\section{Related work}
\label{sec:relatedwork}
Chatbots have become increasingly popular in recent years, as businesses look for ways to automate customer service and improve customer engagement. Generally, chatbots can be categorized into different categories, among them educational, healthcare, and business. They are based on knowledge from those domains to provide relevant support for the user. Educational chatbots are being used to support students, teachers, and administrators, providing personalized learning experiences and improving student outcomes. For example, educational chatbots can provide information on academic subjects, answer questions about the curriculum, and provide feedback on student performance. One example is MedChatbot \cite{Kazi2012MedChatBotAU} for medical students, which is based on open-source AIML. Authors in this chatbot deploy the widely available  Unified Medical Language System (UMLS) as the domain knowledge source for generating responses to queries and converting natural language queries into relevant  SQL  queries. These SQL queries are run against the knowledge base, and results are returned to the user in the natural dialogue. Computer Simulation in Educational Communication, CSIEC \cite{JIA2009249}, is a system with newly developed multiple functions for English instruction that can chat interactively in English with English learners. It generates responses according to the user input, the dialogue context, user and personality knowledge, common sense knowledge, and inference knowledge. Freudbot \cite{Heller2005FreudbotAI} for the psychology domain, is designed to chat in the first person about his theories, concepts, and biographical events using a resource that contained dictionary-type definitions of Freudian terms and concepts. Other chatbots for FAQs have been proposed, among them \cite{qiu-etal-2017-alime}.

Healthcare chatbots are being used to improve the patient experience, reduce wait times, and improve triage. For example, virtual health assistants can provide medical information, schedule appointments, and even help diagnose certain conditions. These chatbots are also being used to improve communication between patients and healthcare providers, making it easier for patients to access the information and support they need. Among those chatbots, Mamabot for supporting women and families during pregnancy \cite{Vaira2018MamaBotAS} exploiting AI-based chatbot systems to understand and respond to the needs of patients and their families. Also, Pharmabot \cite{pharmabot} in 2015 is a pediatric generic medicine consultant chatbot designed to prescribe, suggest, and give information on generic medicines for children. A Chatbot for Psychiatric Counseling in Mental Healthcare Service Based on Emotional Dialogue Analysis and Sentence Generation \cite{Oh2017ACF} that suggests a conversational service for psychiatric counselling that is adapted methodologies to understand counselling contents based on high-level natural language understanding (NLU). Dyva \cite{dyva} proposed a medical chatbot; the idea is to create a text-to-text chatbot that engages patients in conversation about their medical issues and provides basic information and diagnosis based on their symptoms. To improve the quality of life for young adults with food allergies, AllergyBot \cite{Hsu:2017} was proposed as an intelligent and humane chatbot that provides restaurants' allergy accommodation information based on users' allergens.

In addition, chatbots from the business domain are also proposed to support customers and improve companies' services in e-commerce systems. Customer service chatbots have been widely adopted by businesses of all sizes to provide 24/7 support to customers. These chatbots use natural language processing (NLP) and machine learning (ML) algorithms to understand customer inquiries and provide relevant responses. They can be integrated with popular messaging platforms, such as Facebook Messenger and WhatsApp, to make it easy for customers to interact with the chatbot. Among those chatbots, SuperAgent \cite{cui2017superagent} from Microsoft is a customer service chatbot that leverages large-scale and publicly available e-commerce data. It takes advantage of data from in-page product descriptions and user-generated content from e-commerce websites, which is more practical and cost-effective when answering repetitive questions. Another conversational system is from IBM to automatically generates responses for users' requests on social media integrated with state-of-the-art deep learning techniques and trained by nearly 1M Twitter conversations between users and agents from over 60 brands \cite{Xu:2017:NCC:3025453.3025496}. This system adopts a word embedding method, word2vec neural network language model \cite{Mikolov:2013:DRW:2999792.2999959} to learn distributed representations of words from the customer.

Many open-source frameworks that assist in easily creating centralized conversational engines are proposed, among them, Microsoft Bot, Facebook Messenger, Google Assistant, and Amazon Lex. These frameworks build and connect intelligent conversation engines to interact with customers naturally wherever they are by taking advantage of the wide range of users. Besides, they are highly customizable in real scenarios with third-party data. However, most of them are limited regarding privacy, trust, and data governance. There are many situations where data cannot or should not be shared in this way, such as when data is regulated, sensitive or private (medical text, business, social media). An obvious example is biomedical use cases where a single hospital or clinic may not have access to a sufficient amount of data. Thus the performance of predictive models stands in direct conflict with patient privacy.

In this context, we here explore the use of federated learning as a method for distributed and collaborative machine learning to train large models for chatbots and conversational agents. Organizations maintain and govern their data locally and participate in learning a new global, federated model by sending only their model updates (model weights) to a server for aggregation into the global model. Hence, all participants (clients) can benefit from a newly trained model without exposing their data publicly. Our contributions in this paper can be summarized as follows: (1) We propose federated learning models for a customer support chatbot using a Deep Bidirectional Transformer architecture, (2) we design and develop Fedbot, a trustworthy federated learning system for collaborative training (4) we improve the local training with an incremental learning scheme and private online labelling, and (5) our analyses of the models respect data privacy regulations and performs well compared to a baseline centralized model on customer support data after a couple of rounds.

\section{Proposed approach}
\label{sec:approach}
\begin{figure*}
  \includegraphics[width=\linewidth]{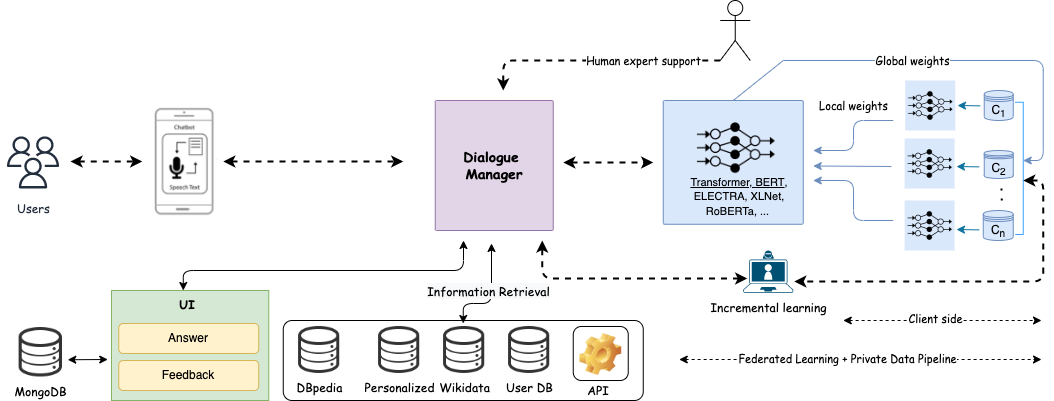}
  \caption{The Fedbot architecture is organized in three main logical layers, the first one deals with privacy-preserving training using federated learning, the second one is a dialogue manager, and the third is for incremental learning and private data labelling.}
  \label{fig1:system}
\end{figure*}

The overall architecture of our proposed Fedbot is shown in Figure \ref{fig1:system}; it is based on a modular approach and takes advantage of scalable federated learning, data privacy, and distributed ML. Fedbot is trained on customer support data on Twitter, (e.g., over 3 million tweets and replies from the biggest brands) and can respond to user queries related to customer support; it can also retrieve information from multiple sources (information retrieval). Users can interact with the Fedbot using a chit-chat system or voice-based messages. The main modules are the private data pipeline, federated learning settings, a dialogue manager, incremental learning, and information retrieval (see Figure \ref{fig1:system}). The private data pipeline module allows parties (clients) to process and prepare their data locally to be used by federated learning {methods}. The federated learning module enables multiple private clients to form an alliance to collaboratively train transformer models and send parameters to the server for global model generation (aggregation of local models). The information retrieval module allows users to enrich the conversation with the chatbot by using open-linked data and knowledge graphs. The system allows parties to add new data continuously and train the model through a defined number of rounds to improve the performance using incremental learning techniques. Finally, participating clients can use the global model {for customer support}. The system is implemented using the open source federated learning framework FEDn ~\cite{ekmefjord2021scalable} and a web interface using Flask (\url{https://flask.palletsprojects.com/} accessed on: 04 Avril 2023). {FEDn provides} a highly scalable federated learning run-time, and Flask is used to develop interactive and user-friendly dialogue interfaces for the different processes in the workflow. The list of available datasets related to customer support used for the demo is placed in the local data sources.
Moreover, the developed system is scalable, flexible, and can be expanded with new clients/data sets on-demand (without re-training the federated model). The conversations and user feedback are stored in a database anonymously for continuous learning and future improvement. More details about the working mechanism of the proposed Fedbot will be explained in the following subsections.

\subsection{Private Data Pipeline Module}
Deep learning has become the main component of many popular conversational AI and chatbot tasks, such as intent classification, speech recognition, natural language understanding, and question-answering. With massive generated data, these tasks require computational and storage resources. To support this increased demand, cloud providers such as Google cloud \cite{googl}, Microsoft Azure \cite{azure}, and Amazon \cite{aws} provide machine learning and deep learning pipelines as a service to train and serve models. Customers must provide/upload training datasets to the cloud provider to use these services. However, training datasets can contain sensitive and private data (e.g., personal medical data, customer data, and financial data). Hence, data privacy becomes a security problem in machine learning with its unique challenges. 
We propose a customer support chatbot based on a federated learning methodology to preserve data privacy and provide secure collaborative training. Fedbot builds on federated learning in which participating clients train local model updates and then share the gradient (model parameters) with the central server. The server implements an  aggregation strategy where local models are combined into an updated global model. This whole process is iterated until convergence. This approach ensures input data privacy, enables collaborative training, and reduces the need for very large central computing resources by distributing the workload across clients. 
Further, the proposed system is capable of improving local learning by using incremental learning and a local data labelling pipeline.

\subsection{Federated Machine Learning {Module}}
Federated learning is an emerging technology for collaborative and decentralized machine learning \cite{DBLP:journals/corr/abs-1912-04977} in which multiple parties (client) jointly train machine learning models using private data. These parties could be mobile and IoT devices (cross-device FL) or larger entities such as different organizations,  subsidiaries or geographically separated data silos (cross-silo FL). In FL, private data remains local at each party at all times, and only the parameter updates are communicated with a server. In our system, we use FL to develop a customer support chatbot based on transformer {architecture}. Fedbot trains a global model using \textit{Federated Averaging} ~\cite{mcmahan2017communicationefficient} (Algorithm \ref{fedavg} ) on large amounts of data from multiple geographically distributed parties. Each party trains a local transformer model on its data (Algorithm \ref{client-update}) and share parameters {$W_t$} with the central server for aggregation.
In the aggregation part, the aggregator (running in the \emph{combiner} in FEDn~\cite{ekmefjord2021scalable}) combines shared parameters and generates a single global model {$M(W_t)$} for each round using 
incremental averaging.
In this way, Fedbot can learn new knowledge from different clients without exposing or sharing their private data.

\vspace{12pt}
\begin{algorithm}[!h]
\DontPrintSemicolon
\SetAlgoLined
\KwIn{$W_t$}
\KwOut{$M(W_t)$}
\BlankLine
\textbf{Server executes:}\\
initialized \textbf{$W_0$}

\SetKwFunction{Server}{IncrementalFedAVG}
\SetKwProg{Fn}{Function}{:}{}
\Fn{\Server{$k, W_{t-1}, W_t$}}{

\ForEach{$t\leftarrow 1$ \KwTo $r$}{

$S_t \leftarrow$ (sample a random set of clients)\\
\ForEach{client $k \in S_t$ \textbf{in parallel}}
{
$W_{t+1}^k \leftarrow ClientUpdate(k, W_t,N_l)$\;
$W_{t+1} \leftarrow \sum_{k=1}^{k} \frac{n_k}{n} W_{t+1}^k$\
}
$W_t \leftarrow (W_{t-1} + (W_{t}-W_{t-1})/t)$\
}
\Return $M(W_t)$

}
\caption{{Incremental FedAVG algorithm. \textbf{k}: Number of clients, \textbf{r}: Number of rounds, $W_i$: Local model weights and \textbf{M}: Global model weights.}}
\label{fedavg}
\end{algorithm}

\vspace{12pt}

\begin{algorithm}[!h]
\DontPrintSemicolon
\SetAlgoLined
\KwOut{\textbf{$W_t$}}

\SetKwFunction{client}{ClientUpdate} // Run on client k\\
\SetKwProg{Fn}{Function}{:}{}
\Fn{\client{$k, W_t$}}{

$\beta$ $\leftarrow $ (split~$D^k$~into~mini~batches)\\
\For{$local~epoch~e_i \in 1, \dots e$}{
\For{batch~b $\in$ $\beta$}{
$W_t \leftarrow W_t - \eta \nabla l(W_t,b)$\;
}
}
\KwRet{$W_t$}\;
}
\caption{Local client update, \textbf{k}: Number of clients, $D^k$: Client k local dataset, \textbf{e}: Number of local epochs, and $\eta$ is the learning rate. }
\label{client-update}
\end{algorithm}

\subsection{Transformer architecture }
\begin{figure}
\centering
  \includegraphics[width=8cm]{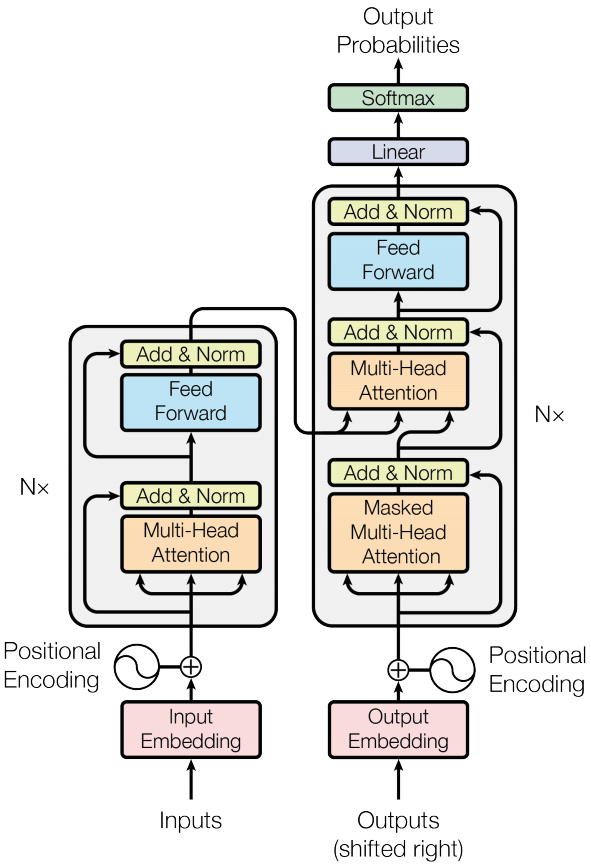}
  \caption{ The Transformer model architecture \cite{vaswani2017attention} - Adapting the transformer model to a Fedbot architecture for natural language processing, featuring an encoder-decoder architecture that generates responses to user inputs using self-attention and feed-forward neural networks}
  \label{transformer} 
  
\end{figure}

\textbf{Encoder-Decoder}: In Fedbot we used the Transformer architecture \cite{vaswani2017attention} to handle client queries in customer support. A transformer is a model architecture eschewing recurrence and instead relying on an attention mechanism to draw global dependencies between input and output; the transformer follows the encoder and decoder architecture using stacked self-attention. The encoder maps an input sequence of symbol representations $(x_1, ..., x_n)$ to a sequence of continuous representations $z = (z_1, ..., z_n)$. The Decoder then generates an output sequence $(y_1, ..., y_m)$ of symbols one element at a time. Both the encoder (left-side) and decoder (right-side) are shown in Figure \ref{transformer}. Both Encoder and Decoder are composed of layers and sub-layers that can be stacked on top of each other multiple times, which $Nx$ describes in Figure \ref{transformer}. The first is a multi-head self-attention mechanism, and the second is a simple, position-wise, fully connected feed-forward network. The inputs and outputs are first embedded into an n-dimensional space.

\textbf{Attention layer}: The attention layer is a key component of the transformer model used in natural language processing tasks \cite{ShenTaoTianyiZhouGuodongLongJingJiangShiruiPan,Parikh2016}. The attention mechanism enables the model to selectively focus on the relevant parts of the input sequence, rather than treating all inputs equally. The attention layer consists of three sets of parameters - query (Q), key (K), and value (V) - that are learned during the training process. In this paper, by applying the self-attention mechanism, we aim at capturing the long dependencies in the input sentence following these steps:
\begin{itemize}
    \item Obtain each word's weight by calculating the similitude between the query matrix and every key matrix $(Q, K, V)$. The dot product, concatenation, and perceptron are typically applied as similarity functions.
    \item Normalize the similitude score of the preceding step and compute the weights with the SoftMax function. 
    \item Finally, the weights and the $V$-values matrix are weighted to get the final attention. Then, the dot product function is used to compute the similarity, and the attention is defined as follows: 
    \begin{equation}
        \begin{aligned}
        Attention(Q,K,V) = softmax(\frac{QK^T}{\sqrt{d_k}})V
        \label{eq3}
        \end{aligned}
    \end{equation}
\end{itemize}
where $QK^T$ represents the dot product between the query vector $Q$ and the transpose of the key matrix $K$, and ${\sqrt{d_k}}$ represents the square root of the dimensionality of the key vectors.\\

Multi-head Attention \cite{vaswani2017attention} is a module for attention mechanisms that run through an attention mechanism several times in parallel. The independent attention outputs are then concatenated and linearly transformed into the expected dimension. The computation of Multi-head Attention proceeds as follows:
\begin{equation}
    \begin{aligned}
    Head_i = Attention(QW_i^Q, KW_i^K  ,VW_i^V)\\
    MultiHead(H') = (Head_i \oplus,..., \oplus Head_h)
    \label{eq3}
    \end{aligned}
\end{equation}

Where the projections are parameter matrices $W_i^Q \in R^{d_{model} d_k}$ and  $W_i^k \in  R^{d_{model}d_k}$ and $W_i^V \in R^{d_{model}d_k}$ and $W_0= R^{(hd_v d_{model})}$.

\subsubsection{Datasets}
\begin{figure*}
  \includegraphics[width=\linewidth]{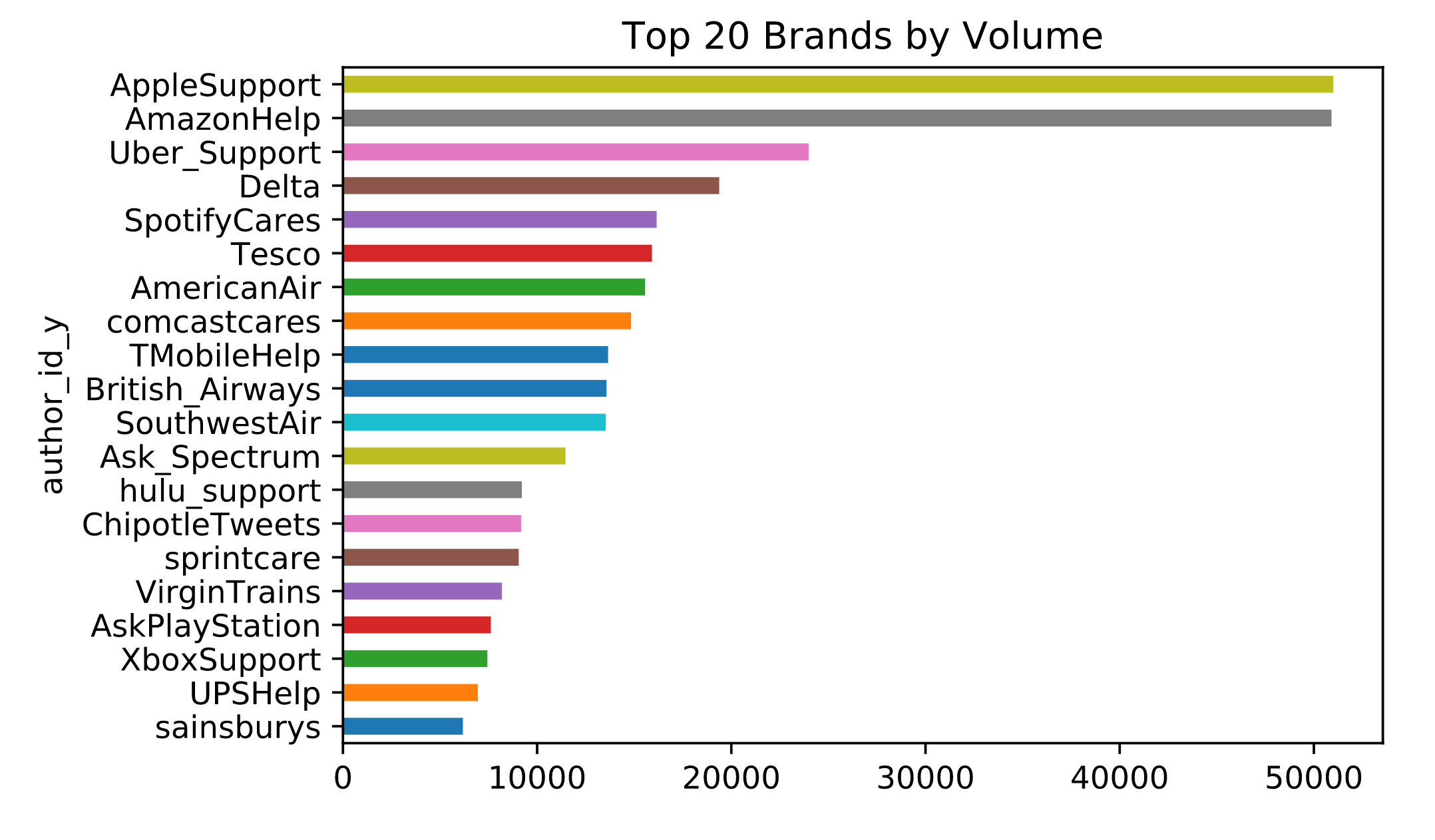}
  \caption{The Customer Support on the Twitter dataset is a large, modern corpus of tweets and replies to aid innovation in natural language understanding. In our Fedbot, each company dataset is considered to be a client in a cross-silo federation that trains a transformer model locally and sends model parameters to a central server for aggregation.}
  \label{dataset} 
\end{figure*}

To train the transformer model, we processed customer support on the Twitter dataset \footnote{https://www.kaggle.com/thoughtvector/customer-support-on-twitter/data} which is a large, modern corpus of tweets and replies to aid innovation in natural language understanding and conversational AI models. The dataset is a CSV file, where each row is a tweet; every conversation included has at least one request from a consumer and at least one response from a company support team. The dataset consists of conversation support for many companies, among them Apple, Amazon, Uber, Delta, Spotify, Tesco, AmericanAir and Spotify (see Figure \ref{dataset}).
In this paper, we used these datasets as private data and processed them to be used to train a transformer model collaboratively in federated learning settings.

\subsection{Information Retrieval (IR)}
With the rapid progress of the semantic web, a huge amount of structured data has become available on the web (web of data). Making these resources available and useful for end-users is one of the main objectives of linked data \cite{lee-ld} and open-domain chatbots. Hence, using IR techniques to access knowledge and enrich user conversations can be very useful to strengthen the Fedbot performances through SPARQL queries on the top of the known knowledge base (DBpedia, Wikidata, etc.). Fedbot uses Named Entity Recognition (NER) and user intent to generate SPARQL queries regarding factoid and recurrent questions to enrich and improve the chatbot's performance.

\subsection{Incremental learning and Feedback}
\begin{figure}
\centering
\includegraphics[width=8cm]{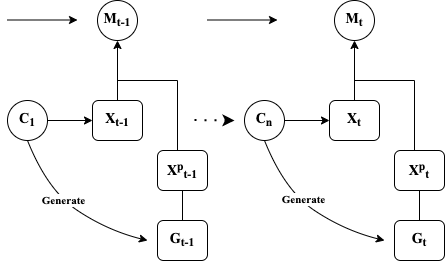}
\caption{Clients can continuously add local private training data to extend the existing model’s knowledge by additional training of the global model. The generated data will be stored locally on the client side. $M_t$ represents the global model. $X_t$ represents the subset of data at time $t$. $G_t$ represents the generator at time $t$, and $X_t^{P}$ is a subset generated by it.}
\label{active_l}
\centering
\end{figure}

In incremental learning, new input data is continuously used to train the model further. A goal is to attempt to improve a model’s performance while adding as few samples as possible. In Fedbot, adding data locally by clients is an important task to improve the local model performance, which is then propagated into the global model after new global training rounds. We have developed an intuitive process for each local client to contribute to the addition of new samples on top of their baseline local data (Figure \ref{active_l}. The first step is to add a new data point that will remain on the local site, allowing users to add private training data. The incremental learning module will process and transform the private data locally and generate training points usable for local model training. This process enables collaborative data generation between organizations while activating data protection and avoiding necessary sharing within the alliance. Furthermore, FedBot utilizes a database layer to store feedback and queries from end users. These inputs are collectively analyzed to enhance and optimize future usage. Additionally, numerous modules and libraries can be readily incorporated with FedBot to broaden its functionalities and effectively handle various tasks.

\section{Experiment and results}
\label{sec:experiments}
We conduct the experiment on the customer support dataset, which contains 550.287 question-answer pairs on 11 datasets from different companies (Figure \ref{dataset}). To ensure collaborative training, we randomly select and split data through 10 clients with 20\% for the validation dataset for all clients. We used an encoder and decoder to build our transformer model, and the implementation is based on TensorFlow and FEDn. For the training process, we set the initial learning rate to 0.001, the batch size to 32 and the maximum number of epochs to 10. Texts are tokenized using Wordpieces \cite{wu2016googles} with a maximum length of 30. More configuration is shown in Table \ref{training-config}. The hyper-parameters used in this model are shown in Table \ref{hyper-params}.

\begin{table}
\centering
  \caption{Federated training configuration}
  \label{training-config}
  \begin{tabular}{lllll}
  \hline\noalign{\smallskip}
    Rounds & Nbr clients & Update size &  Nbr of parameters\\
    \noalign{\smallskip}\hline\noalign{\smallskip}
	30 & 10 & 72 MB & 19.639.619\\ 
\noalign{\smallskip}\hline
\end{tabular}
\end{table}

\begin{table*}
\centering
  \caption{ Order in which different hyper-parameters are explored and the corresponding values considered
for each hyper-parameter. Underlined values indicate the default value.}
  \label{hyper-params}
  \begin{tabular}{llll}
  \hline\noalign{\smallskip}
    Step & Hyper-parameter & Values\\
    \noalign{\smallskip}\hline\noalign{\smallskip}
	1 & embedding dimension & 256, 512, 1024 \\ 
	2 & attention heads & 1, 2, 4, \underline{8}, 16 \\ 
	3 & dropout & 0.1, \underline{0.2}, 0.3, 0.4, 0.5 \\
	4 & number of layers  & 1, 2, 3, \underline{4}, 5, 6, 7, 8 \\
	5 & number of units & 128, 256, \underline{512} \\
	6 & enc/dec layer dropout & 0, 0.1, \underline{0.2}, 0.3, 0.4 \\
	7 & attention dropout  & 0, 0.1, \underline{0.2}, 0.3 \\
	8 & activation dropout & 0, 0.1, \underline{0.2}, 0.3, 0.4, 0.5 \\
	9 & batch size & 8, \underline{32}, 64 \\
	10 & learning rate scheduler & \underline{Transformer standard}, inverse square root \\
	11 & warm-up steps  & 2000, \underline{4000}, 5000, 6000, 8000, 10000 \\
	12 &  learning rate  & 0.01, \underline{0.001}, 0.0001, 0.00001 \\
	13 &  epochs & 5, \underline{10}, 15, 20, 25 \\
	14 &  maxlen & 15, 20, 25, \underline{30} \\
	15 &  activation function & \underline{relu} \\
\noalign{\smallskip}\hline
\end{tabular}
\end{table*}

\subsection{Evaluation}
To evaluate the performance of Fedbot, we used accuracy and loss metrics. We then evaluated performance using the accuracy function provided by Tensorflow that falls between (0, 100), shown in Equation \ref{eq:accuracy}, and the categorical cross-entropy as a loss function, shown in Equation \ref{eq:loss}.

\begin{equation}
 Accuracy= \frac{TP + TN}{TP +TN + FN + FP}*100\%
 \label{eq:accuracy}
\end{equation}

\noindent where $TP$, $TN$,$FP$, and $FN$  are the True Positives, True Negatives, False Positives and False Negatives respectively.
\begin{equation}
Loss= -\sum_{i=1}^{N} y_i.\log\hat{y}_i
\label{eq:loss}
\end{equation}

where $\hat{y}_i$ is the model prediction for \textit{i-th} pattern, $y_i$ represent the corresponding real value, and \textit{N} is the total number of samples.

\begin{figure}[!h]
\begin{minipage}[t]{0.5\linewidth}
    \centering
    \includegraphics[width=1\textwidth]{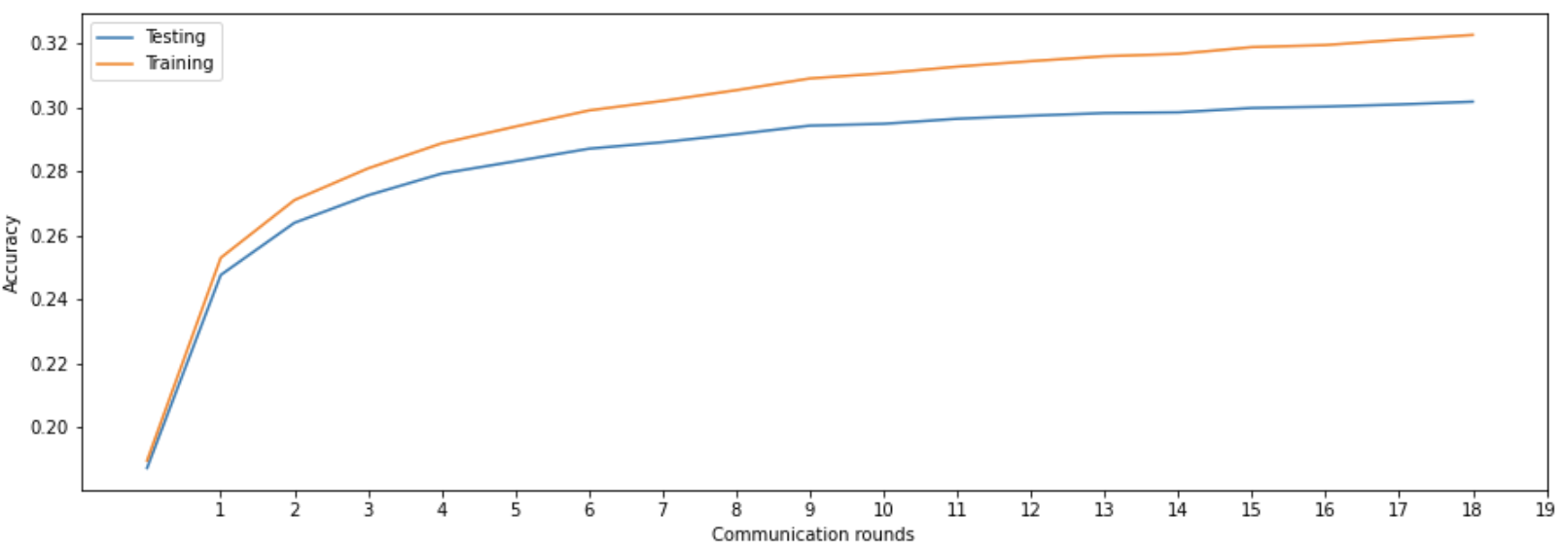}
    \caption{Convergence of training and testing accuracy on customer support dataset for 10 clients on 20 communication rounds}
    \label{fig:convergence}
\end{minipage}
\hspace{0.1cm}
\begin{minipage}[t]{0.5\linewidth} 
    \centering
    \includegraphics[width=1\textwidth]{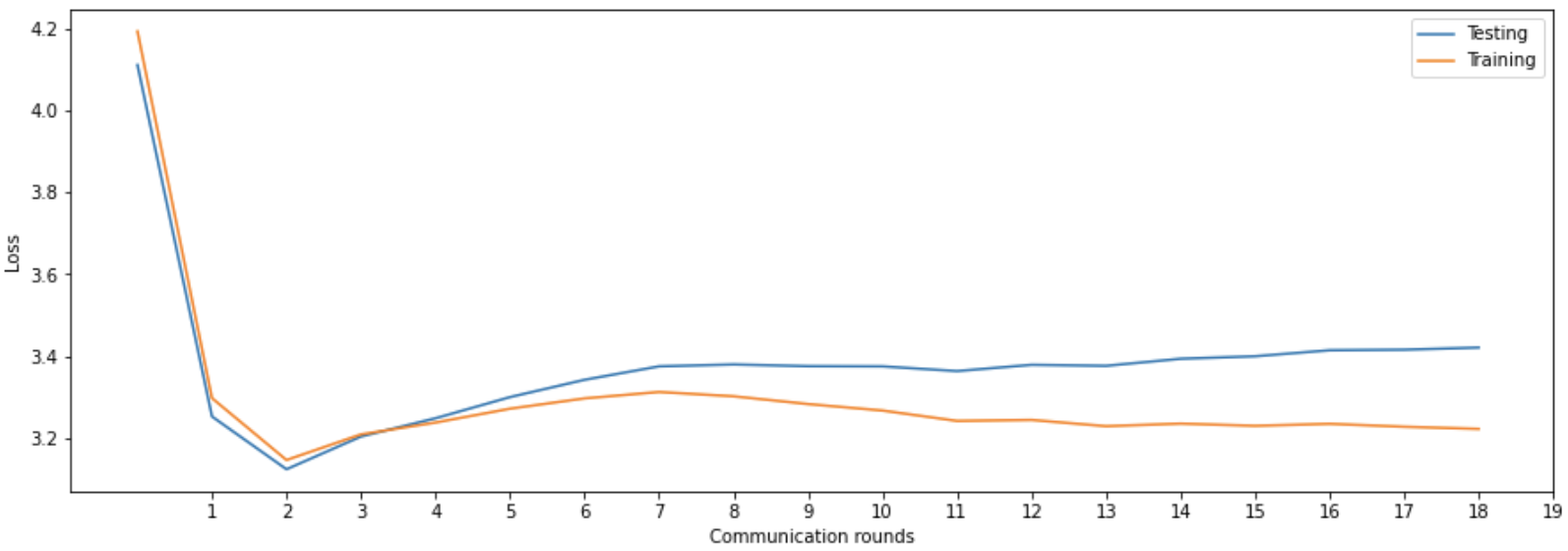}
    \caption{Divergence of training and testing loss for 10 clients on 20 communication rounds.}
    \label{fig:divergence}
\end{minipage}        
\end{figure}

\begin{figure}[!h]
\begin{minipage}[t]{0.5\linewidth}
    \centering
    \includegraphics[width=1\textwidth]{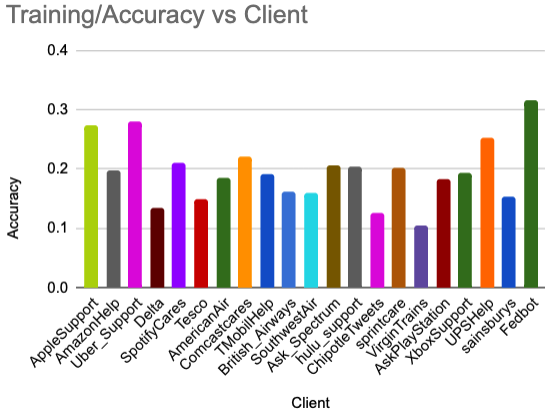}
    \caption{Training accuracy for partial models trained on each customer support dataset. The accuracy varies greatly between clients due to the amount of available data. As can be seen, the federated Fedbot model improves on all client models.}
    \label{fig:training_partial_models}
\end{minipage}
\hspace{0.1cm}
\begin{minipage}[t]{0.5\linewidth} 
    \centering
    \includegraphics[width=1\textwidth]{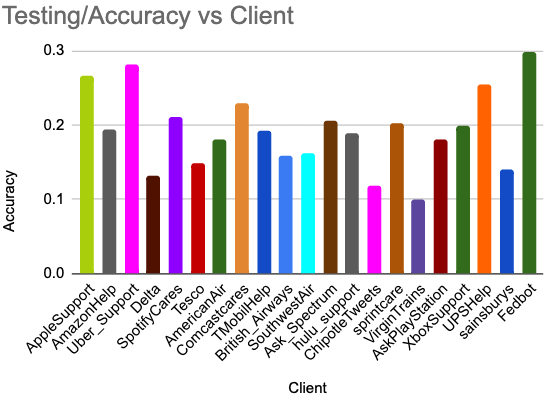}
    \caption{Testing accuracy for partial models on each customer support dataset, resulted in a difference due to the amount of available data. In the end, a partial model (client) with poor performance will benefit from knowledge transfer via the global model}
    \label{fig:testing_partial_models}
\end{minipage}        
\end{figure}

To validate our POC, we partitioned the dataset into ten equal chunks so that each client has 20\% of the total dataset. We then compare the federated scenario to centralized model training using the entire dataset. The convergence and divergence of training and testing accuracy on the customer support dataset for 10 clients on 20 communication rounds are depicted in Figure \ref{fig:convergence} and Figure \ref{fig:divergence}, respectively. In addition Figure, \ref{fig:training_partial_models} and Figure \ref{fig:testing_partial_models} show the partial models compared to the global model (Fedbot). Partial models represent the model updates contributed by individual clients participating in the collaborative learning process. These updates capture the local insights and unique data characteristics of each client, which are incorporated into the global model to enhance its performance. The use of partial models in federated learning highlights the idea that the model's performance benefits everyone involved in the collaborative learning process. Each device contributes its knowledge to the shared model, which improves the overall model's accuracy, and this, in turn, benefits all the participants.

For demonstration, we consider a FEDn network consisting of a single, high-powered aggregation server (known as a combiner) with 16 VCPU and 16GB RAM, along with connected client instances (each with 16 VCPU and 16GB RAM) located in SSC (SNIC Science Cloud \cite{8109140}). To have an intuitive observation of the predictions, we give prediction examples using global models in Figure \ref{use-cases}, which shows the Fedbot performance on a given utterance. Hence, the proposed approach has contributed overall to a better understanding of text conversation, preserving data privacy, and contributing to low-cost training as well as collaborative training using large models.

\subsection{Implementation and demo environment}
Designing and developing privacy-aware chatbots is not a trivial task; it requires various design techniques for data governance and privacy regulations. Third-party frameworks have been proposed for chatbot development; it refers to open-source building blocks that help build conversation engines, including Microsoft Bot, Facebook Messenger, Google Assistant, and Amazon Lex. However, the increasing use of chatbot technologies has led to a heightened focus on privacy issues. To tackle this concern, we developed Febot, a system that employs natural language processing techniques and federated learning to operate on private data. It includes the following benefits:

\begin{itemize}
    \item Privacy Protection: Federated learning ensures that sensitive customer data remains on the local device, reducing the risk of data breaches and providing stronger privacy protection for customers.
    \item Improved User Experience: By training models locally, federated learning enables chatbots to provide more accurate and relevant responses to users. This can result in a better customer experience and higher customer satisfaction.
    \item Scalability: Federated learning enables organizations to scale their chatbots quickly and efficiently, as the models can be trained on multiple devices simultaneously. This reduces the need for large amounts of centralized computing resources and eliminates the need for complex data transfers.
    \item Robust Models: Federated learning enables chatbots to leverage data from multiple devices to train robust models, improving their accuracy and performance.
    \item Incremental learning: generate new knowledge and a new global model by attaching more clients and annotating new data points;
    \item Standalone: multiple platforms (i.e., guarantee for low disk and memory footprint). It can be run on a standard laptop having two cores and 2GB of RAM;
\end{itemize}

\begin{figure*}
  \includegraphics[width=\linewidth]{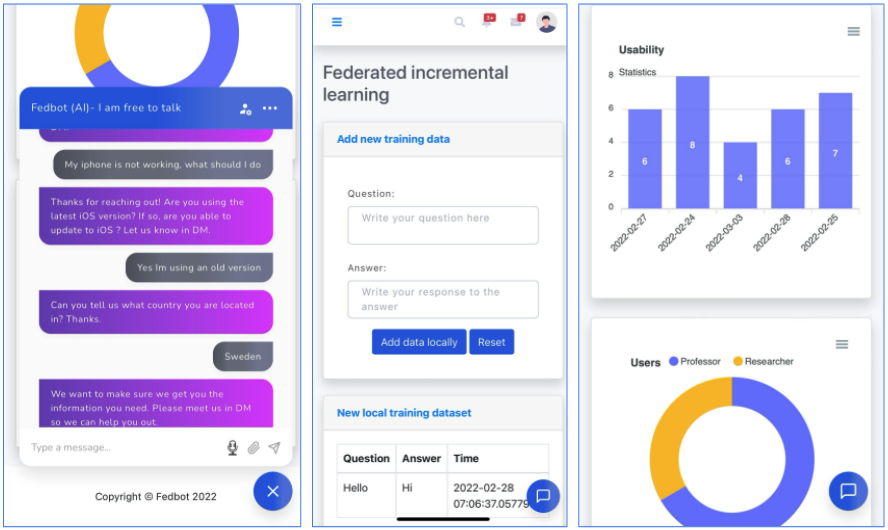}
  \caption{Fedbot use case including UI and federated incremental learning where the model is trained incrementally on new data as it becomes available. This approach allows the model to adapt to changes in the data over time, improving its accuracy and relevance. }
  \label{use-cases} 
\end{figure*}

\vspace{0.2cm}
The proposed Fedbot is accessible from different platforms to engage a wide range of users, and it is also optimized for both desktop and mobile. Figure \ref{use-cases} present some examples.

\section*{Conclusion}
\label{sec:conclusion}
In this paper, we have proposed Fedbot, a proof of concept customer support chatbot system, to address the data-sharing issue in machine learning and to handle client questions in customer support. Validation experiments were implemented based on twenty training rounds with a transformer neural network. The system consists of several components: the private data pipeline, dialogue management, collaborative training, and private incremental learning. Experiments on the customer support dataset using the transformer architecture demonstrate that our approach performs well compared to the baseline centralized model. The proposed Fedbot allows collaborative training participants to control their sensitive and private data while training a chatbot. The integration of federated learning within chatbot and conversation AI provides a sustainable solution by preserving data privacy. We conclude that the application of FL to NLP tasks such as dialogue understanding can contribute to solving the problem that arises when using machine learning with private and sensitive data.
Furthermore, the system actively supports end-users in joining training and improving their performance through incremental learning on various local clients.\\ In future work, we aim to extend Fedbot to cover more customer support datasets in a real geographically distributed manner and test the model with other aggregation algorithms (e.g., FedOPT, FedProx, etc.) for federated learning. Additionally, we intend to Fine-tune the pre-trained model LLaMA \cite{touvron2023llama} and investigate its efficacy in federated learning settings.

\section*{Acknowledgement}
This work is funded by Uppsala University in Sweden on scalable federated learning research and supported by the University of Sk\"ovde. The authors also would like to thank SNIC for providing cloud resources.

\bibliographystyle{unsrtnat}

\bibliography{fedbot}

\begin{thebibliography}{28}
\providecommand{\natexlab}[1]{#1}
\providecommand{\url}[1]{\texttt{#1}}
\expandafter\ifx\csname urlstyle\endcsname\relax
  \providecommand{\doi}[1]{doi: #1}\else
  \providecommand{\doi}{doi: \begingroup \urlstyle{rm}\Url}\fi

\bibitem[Weizenbaum(1966)]{Weizenbaum:1966:ECP:365153.365168}
Joseph Weizenbaum.
\newblock Eliza a computer program for the study of natural language
  communication between man and machine.
\newblock \emph{Commun. ACM}, 9\penalty0 (1):\penalty0 36--45, January 1966.
\newblock ISSN 0001-0782.
\newblock \doi{10.1145/365153.365168}.
\newblock URL \url{http://doi.acm.org/10.1145/365153.365168}.

\bibitem[Colby(1999)]{Colby1999}
K.~M. Colby.
\newblock \emph{Human-Computer Conversation in A Cognitive Therapy Program},
  pages 9--19.
\newblock Springer US, Boston, MA, 1999.
\newblock ISBN 978-1-4757-5687-6.
\newblock \doi{10.1007/978-1-4757-5687-6_3}.
\newblock URL \url{https://doi.org/10.1007/978-1-4757-5687-6_3}.

\bibitem[AbuShawar and Atwell(2015)]{alice001}
Bayan AbuShawar and Eric Atwell.
\newblock Alice chatbot: Trials and outputs.
\newblock \emph{Computación y Sistemas}, 19, 12 2015.
\newblock \doi{10.13053/cys-19-4-2326}.

\bibitem[Ekmefjord et~al.(2021)Ekmefjord, Ait-Mlouk, Alawadi, Åkesson,
  Stoyanova, Spjuth, Toor, and Hellander]{ekmefjord2021scalable}
Morgan Ekmefjord, Addi Ait-Mlouk, Sadi Alawadi, Mattias Åkesson, Desislava
  Stoyanova, Ola Spjuth, Salman Toor, and Andreas Hellander.
\newblock Scalable federated machine learning with fedn, 2021.

\bibitem[Kazi et~al.(2012)Kazi, Chowdhry, and Memon]{Kazi2012MedChatBotAU}
Hameedullah Kazi, Bhawani~Shankar Chowdhry, and Zeesha Memon.
\newblock Medchatbot: An umls based chatbot for medical students.
\newblock 2012.

\bibitem[Jia(2009)]{JIA2009249}
Jiyou Jia.
\newblock Csiec: A computer assisted english learning chatbot based on textual
  knowledge and reasoning.
\newblock \emph{Knowledge-Based Systems}, 22\penalty0 (4):\penalty0 249 -- 255,
  2009.
\newblock ISSN 0950-7051.
\newblock \doi{https://doi.org/10.1016/j.knosys.2008.09.001}.
\newblock URL
  \url{http://www.sciencedirect.com/science/article/pii/S0950705109000045}.
\newblock Artificial Intelligence (AI) in Blended Learning.

\bibitem[Heller et~al.(2005)Heller, Proctor, Mah, Jewell, and
  Cheung]{Heller2005FreudbotAI}
Bob Heller, Mike~D. Proctor, Dean Y.-O.-N. Mah, Lisa Jewell, and Bill Cheung.
\newblock Freudbot: An investigation of chatbot technology in distance
  education.
\newblock 2005.

\bibitem[Qiu et~al.(2017)Qiu, Li, Wang, Gao, Chen, Zhao, Chen, Huang, and
  Chu]{qiu-etal-2017-alime}
Minghui Qiu, Feng-Lin Li, Siyu Wang, Xing Gao, Yan Chen, Weipeng Zhao, Haiqing
  Chen, Jun Huang, and Wei Chu.
\newblock {A}li{M}e chat: A sequence to sequence and rerank based chatbot
  engine.
\newblock In \emph{Proceedings of the 55th Annual Meeting of the Association
  for Computational Linguistics (Volume 2: Short Papers)}, pages 498--503,
  Vancouver, Canada, July 2017. Association for Computational Linguistics.
\newblock \doi{10.18653/v1/P17-2079}.
\newblock URL \url{https://www.aclweb.org/anthology/P17-2079}.

\bibitem[Vaira et~al.(2018)Vaira, Bochicchio, Conte, Casaluci, and
  Melpignano]{Vaira2018MamaBotAS}
Lucia Vaira, Mario~A. Bochicchio, Matteo Conte, Francesco~Margiotta Casaluci,
  and Antonio Melpignano.
\newblock Mamabot: a system based on ml and nlp for supporting women and
  families during pregnancy.
\newblock In \emph{IDEAS}, 2018.

\bibitem[Comendador et~al.(2015)Comendador, Francisco, Medenilla, Nacion, and
  Serac]{pharmabot}
Benilda~Eleonor Comendador, Bien Francisco, Jefferson Medenilla, Sharleen
  Nacion, and Timothy Serac.
\newblock Pharmabot: A pediatric generic medicine consultant chatbot.
\newblock \emph{Journal of Automation and Control Engineering}, 3:\penalty0
  137--140, 01 2015.
\newblock \doi{10.12720/joace.3.2.137-140}.

\bibitem[Oh et~al.(2017)Oh, Lee, Ko, and Choi]{Oh2017ACF}
KyoJoong Oh, Dongkun Lee, ByungSoo Ko, and Ho-Jin Choi.
\newblock A chatbot for psychiatric counseling in mental healthcare service
  based on emotional dialogue analysis and sentence generation.
\newblock \emph{2017 18th IEEE International Conference on Mobile Data
  Management (MDM)}, pages 371--375, 2017.

\bibitem[Divya et~al.()Divya, Indumathi, Ishwarya, Priyasankari, and
  Devi]{dyva}
Divya, Indumathi, Ishwarya, Priyasankari, and Kalpana Devi.
\newblock A self-diagnosis medical chatbot using artificial intelligence.
\newblock \emph{Journal of Web Development and Web Designing}, 3.

\bibitem[Hsu et~al.(2017)Hsu, Zhao, Liao, Liu, and Wang]{Hsu:2017}
Paris (Pei-Ting) Hsu, Jingshu Zhao, Kehan Liao, Tianyi Liu, and Chen Wang.
\newblock Allergybot: A chatbot technology intervention for young adults with
  food allergies dining out.
\newblock In \emph{Proceedings of the 2017 CHI Conference Extended Abstracts on
  Human Factors in Computing Systems}, CHI EA '17, pages 74--79, New York, NY,
  USA, 2017. ACM.
\newblock ISBN 978-1-4503-4656-6.
\newblock \doi{10.1145/3027063.3049270}.
\newblock URL \url{http://doi.acm.org/10.1145/3027063.3049270}.

\bibitem[Cui et~al.(2017)Cui, Wei, Huang, Tan, Duan, and
  Zhou]{cui2017superagent}
Lei Cui, Furu Wei, Shaohan Huang, Chuanqi Tan, Chaoqun Duan, and Ming Zhou.
\newblock Superagent: A customer service chatbot for e-commerce websites.
\newblock In \emph{Proceedings of ACL 2017, System Demonstrations}, pages
  97--102. Association for Computational Linguistics, July 2017.
\newblock URL
  \url{https://www.microsoft.com/en-us/research/publication/superagent-customer-service-chatbot-e-commerce-websites/}.

\bibitem[Xu et~al.(2017)Xu, Liu, Guo, Sinha, and
  Akkiraju]{Xu:2017:NCC:3025453.3025496}
Anbang Xu, Zhe Liu, Yufan Guo, Vibha Sinha, and Rama Akkiraju.
\newblock A new chatbot for customer service on social media.
\newblock In \emph{Proceedings of the 2017 CHI Conference on Human Factors in
  Computing Systems}, CHI '17, pages 3506--3510, New York, NY, USA, 2017. ACM.
\newblock ISBN 978-1-4503-4655-9.
\newblock \doi{10.1145/3025453.3025496}.
\newblock URL \url{http://doi.acm.org/10.1145/3025453.3025496}.

\bibitem[Mikolov et~al.(2013)Mikolov, Sutskever, Chen, Corrado, and
  Dean]{Mikolov:2013:DRW:2999792.2999959}
Tomas Mikolov, Ilya Sutskever, Kai Chen, Greg Corrado, and Jeffrey Dean.
\newblock Distributed representations of words and phrases and their
  compositionality.
\newblock In \emph{Proceedings of the 26th International Conference on Neural
  Information Processing Systems - Volume 2}, NIPS'13, pages 3111--3119, USA,
  2013. Curran Associates Inc.
\newblock URL \url{http://dl.acm.org/citation.cfm?id=2999792.2999959}.

\bibitem[cloud(2021{\natexlab{a}})]{googl}
Google cloud.
\newblock Google cloud, 2021{\natexlab{a}}.
\newblock URL \url{https://cloud.google.com/}.

\bibitem[cloud(2021{\natexlab{b}})]{azure}
Azure cloud.
\newblock Azure cloud, 2021{\natexlab{b}}.
\newblock URL \url{https://azure.microsoft.com/en-us/}.

\bibitem[cloud(2021{\natexlab{c}})]{aws}
AWS cloud.
\newblock Aws cloud, 2021{\natexlab{c}}.
\newblock URL \url{https://aws.amazon.com/}.

\bibitem[Kairouz et~al.(2019)Kairouz, McMahan, Avent, Bellet, Bennis, Bhagoji,
  Bonawitz, Charles, Cormode, Cummings, D'Oliveira, Rouayheb, Evans, Gardner,
  Garrett, Gasc{\'{o}}n, Ghazi, Gibbons, Gruteser, Harchaoui, He, He, Huo,
  Hutchinson, Hsu, Jaggi, Javidi, Joshi, Khodak, Kone{\v{c}}n{\'y}, Korolova,
  Koushanfar, Koyejo, Lepoint, Liu, Mittal, Mohri, Nock, {\"{O}}zg{\"{u}}r,
  Pagh, Raykova, Qi, Ramage, Raskar, Song, Song, Stich, Sun, Suresh,
  Tram{\`{e}}r, Vepakomma, Wang, Xiong, Xu, Yang, Yu, Yu, and
  Zhao]{DBLP:journals/corr/abs-1912-04977}
Peter Kairouz, H.~Brendan McMahan, Brendan Avent, Aur{\'{e}}lien Bellet, Mehdi
  Bennis, Arjun~Nitin Bhagoji, Keith Bonawitz, Zachary Charles, Graham Cormode,
  Rachel Cummings, Rafael G.~L. D'Oliveira, Salim~El Rouayheb, David Evans,
  Josh Gardner, Zachary Garrett, Adri{\`{a}} Gasc{\'{o}}n, Badih Ghazi,
  Phillip~B. Gibbons, Marco Gruteser, Za{\"{\i}}d Harchaoui, Chaoyang He, Lie
  He, Zhouyuan Huo, Ben Hutchinson, Justin Hsu, Martin Jaggi, Tara Javidi,
  Gauri Joshi, Mikhail Khodak, Jakub Kone{\v{c}}n{\'y}, Aleksandra Korolova,
  Farinaz Koushanfar, Sanmi Koyejo, Tancr{\`{e}}de Lepoint, Yang Liu, Prateek
  Mittal, Mehryar Mohri, Richard Nock, Ayfer {\"{O}}zg{\"{u}}r, Rasmus Pagh,
  Mariana Raykova, Hang Qi, Daniel Ramage, Ramesh Raskar, Dawn Song, Weikang
  Song, Sebastian~U. Stich, Ziteng Sun, Ananda~Theertha Suresh, Florian
  Tram{\`{e}}r, Praneeth Vepakomma, Jianyu Wang, Li~Xiong, Zheng Xu, Qiang
  Yang, Felix~X. Yu, Han Yu, and Sen Zhao.
\newblock Advances and open problems in federated learning.
\newblock \emph{CoRR}, abs/1912.04977, 2019.
\newblock URL \url{http://arxiv.org/abs/1912.04977}.

\bibitem[McMahan et~al.(2017)McMahan, Moore, Ramage, Hampson, and
  y~Arcas]{mcmahan2017communicationefficient}
H.~Brendan McMahan, Eider Moore, Daniel Ramage, Seth Hampson, and
  Blaise~Agüera y~Arcas.
\newblock Communication-efficient learning of deep networks from decentralized
  data, 2017.

\bibitem[Vaswani et~al.(2017)Vaswani, Shazeer, Parmar, Uszkoreit, Jones, Gomez,
  Kaiser, and Polosukhin]{vaswani2017attention}
Ashish Vaswani, Noam Shazeer, Niki Parmar, Jakob Uszkoreit, Llion Jones,
  Aidan~N. Gomez, Lukasz Kaiser, and Illia Polosukhin.
\newblock Attention is all you need, 2017.

\bibitem[Shen et~al.(2018)Shen, Jiang, Zhou, Pan, Long, and
  Zhang]{ShenTaoTianyiZhouGuodongLongJingJiangShiruiPan}
Tao Shen, Jing Jiang, Tianyi Zhou, Shirui Pan, Guodong Long, and Chengqi Zhang.
\newblock {Disan: Directional self-attention network for RnN/CNN-free language
  understanding}.
\newblock In \emph{32nd AAAI Conference on Artificial Intelligence, AAAI 2018},
  pages 5446--5455, 2018.
\newblock ISBN 9781577358008.

\bibitem[Parikh et~al.(2016)Parikh, T{\"{a}}ckstr{\"{o}}m, Das, and
  Uszkoreit]{Parikh2016}
Ankur~P Parikh, Oscar T{\"{a}}ckstr{\"{o}}m, Dipanjan Das, and Jakob Uszkoreit.
\newblock {A decomposable attention model for natural language inference}.
\newblock In \emph{EMNLP 2016 - Conference on Empirical Methods in Natural
  Language Processing, Proceedings}, pages 2249--2255, 2016.
\newblock ISBN 9781945626258.
\newblock \doi{10.18653/v1/d16-1244}.

\bibitem[Berners-Lee(2009)]{lee-ld}
Tim Berners-Lee.
\newblock https://www.w3.org/designissues/linkeddata.html, 2009.

\bibitem[Wu et~al.(2016)Wu, Schuster, Chen, Le, Norouzi, Macherey, Krikun, Cao,
  Gao, Macherey, Klingner, Shah, Johnson, Liu, Łukasz Kaiser, Gouws, Kato,
  Kudo, Kazawa, Stevens, Kurian, Patil, Wang, Young, Smith, Riesa, Rudnick,
  Vinyals, Corrado, Hughes, and Dean]{wu2016googles}
Yonghui Wu, Mike Schuster, Zhifeng Chen, Quoc~V. Le, Mohammad Norouzi, Wolfgang
  Macherey, Maxim Krikun, Yuan Cao, Qin Gao, Klaus Macherey, Jeff Klingner,
  Apurva Shah, Melvin Johnson, Xiaobing Liu, Łukasz Kaiser, Stephan Gouws,
  Yoshikiyo Kato, Taku Kudo, Hideto Kazawa, Keith Stevens, George Kurian,
  Nishant Patil, Wei Wang, Cliff Young, Jason Smith, Jason Riesa, Alex Rudnick,
  Oriol Vinyals, Greg Corrado, Macduff Hughes, and Jeffrey Dean.
\newblock Google's neural machine translation system: Bridging the gap between
  human and machine translation, 2016.

\bibitem[Toor et~al.(2017)Toor, Lindberg, Falman, Vallin, Mohill, Freyhult,
  Nilsson, Agback, Viklund, Zazzik, Spjuth, Capuccini, Möller, Murtagh, and
  Hellander]{8109140}
Salman Toor, Mathias Lindberg, Ingemar Falman, Andreas Vallin, Olof Mohill,
  Pontus Freyhult, Linus Nilsson, Martin Agback, Lars Viklund, Henric Zazzik,
  Ola Spjuth, Marco Capuccini, Joakim Möller, Donal Murtagh, and Andreas
  Hellander.
\newblock Snic science cloud (ssc): A national-scale cloud infrastructure for
  swedish academia.
\newblock In \emph{2017 IEEE 13th International Conference on e-Science
  (e-Science)}, pages 219--227, 2017.
\newblock \doi{10.1109/eScience.2017.35}.

\bibitem[Touvron et~al.(2023)Touvron, Lavril, Izacard, Martinet, Lachaux,
  Lacroix, Rozière, Goyal, Hambro, Azhar, Rodriguez, Joulin, Grave, and
  Lample]{touvron2023llama}
Hugo Touvron, Thibaut Lavril, Gautier Izacard, Xavier Martinet, Marie-Anne
  Lachaux, Timothée Lacroix, Baptiste Rozière, Naman Goyal, Eric Hambro,
  Faisal Azhar, Aurelien Rodriguez, Armand Joulin, Edouard Grave, and Guillaume
  Lample.
\newblock Llama: Open and efficient foundation language models, 2023.

\end{thebibliography}
\end{document}